\documentclass[letterpaper, 10 pt, conference]{ieeeconf}

\IEEEoverridecommandlockouts                              

\usepackage{graphicx} 
\usepackage{amsmath} 
\usepackage{balance}
\usepackage{algorithm,algorithmic}

\title{\LARGE \bf
Radar-Only Off-Road Local Navigation
}

\author{Timothy Overbye and Srikanth Saripalli
\thanks{Timothy Overbye and Srikanth Saripalli are with the Department of Mechanical Engineering, Texas A\&M University, College Station, TX 77840, USA
        {\tt\small overbye2@tamu.edu}
        {\tt\small ssaripalli@tamu.edu}}%
}

\begin{document}

\maketitle
\thispagestyle{empty}
\pagestyle{empty}

\begin{abstract}

Off-road robotics have traditionally utilized lidar for local navigation due to its accuracy and high resolution. However, the limitations of lidar, such as reduced performance in harsh environmental conditions and limited range, have prompted the exploration of alternative sensing technologies. This paper investigates the potential of radar for off-road local navigation, as it offers the advantages of a longer range and the ability to penetrate dust and light vegetation. We adapt existing lidar-based methods for radar and evaluate the performance in comparison to lidar under various off-road conditions. We show that radar can provide a significant range advantage over lidar while maintaining accuracy for both ground plane estimation and obstacle detection. And finally, we demonstrate successful autonomous navigation at a speed of 2.5~m/s over a path length of 350~m using only radar for ground plane estimation and obstacle detection.

\end{abstract}

\section{INTRODUCTION}

Off-road robotics have emerged as a significant area of research in recent years. Perception of the off-road environment is one of the greatest challenges to successful autonomous operation due to its unstructured nature and complexity. Additionally, real world operation should also consider environmental effects such as dust and weather such as rain and fog. For safe and effective operation the perception system should be able to identify terrain features such as slope and roughness, and possible obstacles such as trees, other vehicles, dense vegetation, etc. 

Traditionally lidar and vision have been the primary sensors used in off-road robotics for local navigation. Lidar offers high-resolution spatial data that enables robots to perceive and navigate their surroundings. And vision is capable of providing dense semantic information. However, both lidar and vision have inherent limitations that hinder their effectiveness in certain off-road scenarios. For instance, the performance of both lidar and vision can be significantly affected by adverse environmental conditions such as dust, fog, and rain. Lidar is also not capable of providing the same resolution as vision while vision is still unable to provide accurate depth information. 

As a result, there is a growing interest in exploring alternative sensing technologies that can complement existing sensors in off-road robotics. In this paper we will be looking at radar sensors, which offers potential advantages over lidar in specific situations. Radar can provide longer range sensing capabilities, making it particularly useful for higher speed operation where extended perception range is crucial. Furthermore, radar has the ability to penetrate dust and light vegetation, which can be beneficial in off-road environments. 

In this paper, we focus on off-road local navigation and assess the performance of radar in comparison to lidar. By adapting methods used for lidar we show that off-road navigation is possible with radar alone.

   \begin{figure}[tbp]
      \centering
      \framebox{\parbox{3in}{
      
      \includegraphics[width=\linewidth]{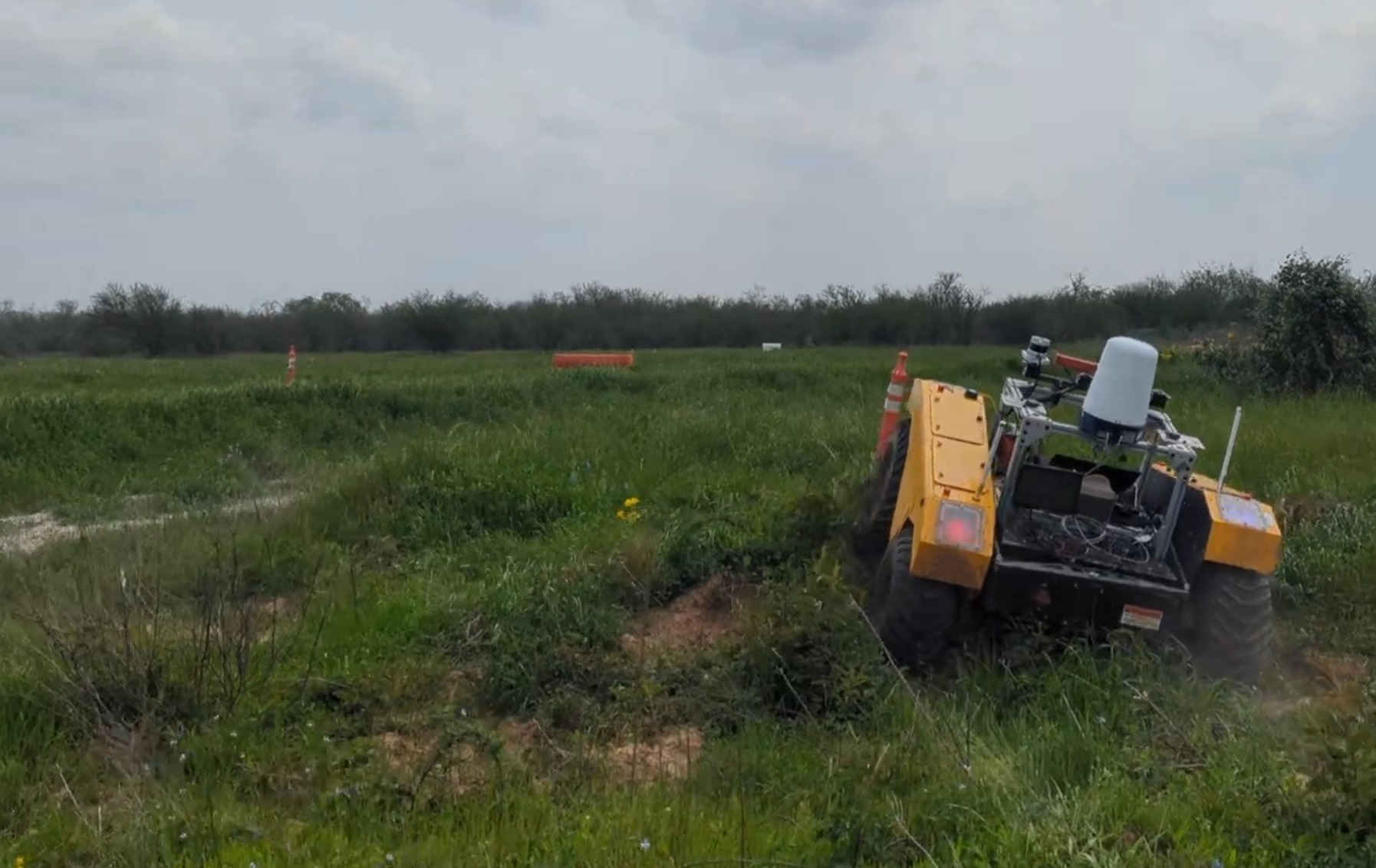}
	}}
      
      \caption{Our vehicle navigating an off-road trail based on a radar map. }
      \label{fig:first_page}
   \end{figure}

\section{RELATED WORK}




There has been a lot of work on radar SLAM, using radar to detect hidden obstacles, and the performance of radar in difficult environments such as smoke, rain, and snow. There has also been some work looking at learning based approaches to obstacle identification using radar. Radar has also been used to supplement various other sensors. However there has not been significant work using radar as the only sensor for autonomous navigation. 

Much of the recent work with radar for autonomous systems has been on the use of radar for Simultaneous Localization and Mapping (SLAM). The work by \cite{radar_slam_2013} presents two distinct approaches for radar-based SLAM. One approach employs the Fourier-Mellin transform for registering radar images in a sequence, while the other capitalizes on movement distortions in data collected from a rotating range sensor to achieve localization and mapping. The authors demonstrate the effectiveness of both methods on real-world data. 

Many of these studies have also looked at a comparison of radar and lidar for SLAM. The work by \cite{radar_lidar_slam_2019} presents a principled comparison of the accuracy of a novel radar sensor against that of a Velodyne lidar, for localization and mapping. The study by \cite{radar_vs_lidar_2009} presents a comparative evaluation of millimeter-wave radar and two-dimensional scanning lasers in dusty and rainy conditions, assessing sensor performance and their effects on terrain mapping and localization. \cite{radar_lidar_slam_2018} explored the fusion of lidar and radar data for SLAM in harsh environments. They show that while radar alone can achieve good results in fog, lidar alone struggles. Although better results are achieved with a fusion of radar and lidar. A more recent study by \cite{radar_slam_2022} extends the radar SLAM topic by comparing three  localization systems: radar-only, lidar-only, and a cross-modal radar-to-lidar system across varying seasonal and weather conditions. their comparison shows that, with modern algorithms, lidar localization may not preform as poorly in bad weather as other studies have shown. However, they do find that radar localization can achieve competitive accuracy to lidar with a much smaller map. 

Recent work has also produced good results by taking existing high resolution lidar maps and using radar to localize the vehicle on these maps~\cite{radar_on_lidar_slam_2020}\cite{radar_on_lidar_slam_2022}. These methods, while primarily of benefit to on road use cases, allow the use of preexisting lidar maps rather than having to create new radar maps for radar localization. 

Also of interest is recent work on obstacle detection using radar. This is of particular interest for off-road robotics due to radar's ability to penetrate some types of vegetation as well as dust and rain. Earlier work \cite{radar_lidar_dust_2009} presents the use of radar as a redundent method of sensing especially under dusty conditions where lidar preforms poorly. This work is continued in \cite{radar_perception_2011} where radar is used identify the ground in low visibility conditions such as dust or looking into the sun and dusk or dawn. Their methods produce good results in situations where lidar or vision might be blinded but are limited to short range sensing. The work in \cite{imaging_radar_2019} presents initial experimental results of radar's ability to detect obstacles obscured by vegetation, enhancing the perception system's capabilities. Another work, \cite{radar_vision_obs_2015} details two strategies for terrain traversability assessment: one using stereo data and the other using an integrated radar-stereo system to detect and characterize obstacles, showing the usefulness of radar in identifying and assessing obstacles in outdoor environments. A similar problem is presented in \cite{radar_vision_obs_2015} where radar and stereo vision are used to both estimate the ground plane and detect obstacles. In this work radar is primarily used for obstacle detection with the stereo camera using a machine learning algorithm to estimate the ground plane. The work in \cite{radar_lidar_dust_2012} compares the quality of lidar and radar for obstacle detection in a dusty off-road environment. Here the radar is proposed to be used as a backup system with the lidar handling most of the work. They find that the lidar produces better results but is blocked by dust, when this happens the radar can take over. Finally, the work in \cite{radar_lidar_obs_2015} demonstrates the benefits associated with lidar and radar sensor fusion, particularly in detecting objects partially obscured by light to medium vegetation.

In this paper we look at a geometric approach based off our previous work G-VOM \cite{g-vom}, an open source lidar mapping package, to use radar alone for off-road navigation. Despite the differences in the sensors types, this can still achieve good results. We also discuss some of the advantages that our method provides over lidar only navigation. 


\section{APPROACH}

\subsection{Sensors}

Before we discuss the differences between radar and lidar it is worth taking a brief look at how they both work and the differences in the returned data. Both technologies rely on time of flight measurements of electromagnetic radiation. Lidar typically functions within the infrared section of the spectrum, while radar operates at higher wavelengths, within the radio band of the spectrum at 77--81~GHz. This attribute allows radar to penetrate some objects that are opaque to lidar. Such as environmental effects including dust and rain, and low density obstacles such as vegetation. However, more important for this work is the difference in the shape of the emitted beam and how the returns are processed into data. 

\begin{figure}[htbp]
      \centering
      \framebox{\parbox{3in}{
      
      \includegraphics[width=\linewidth]{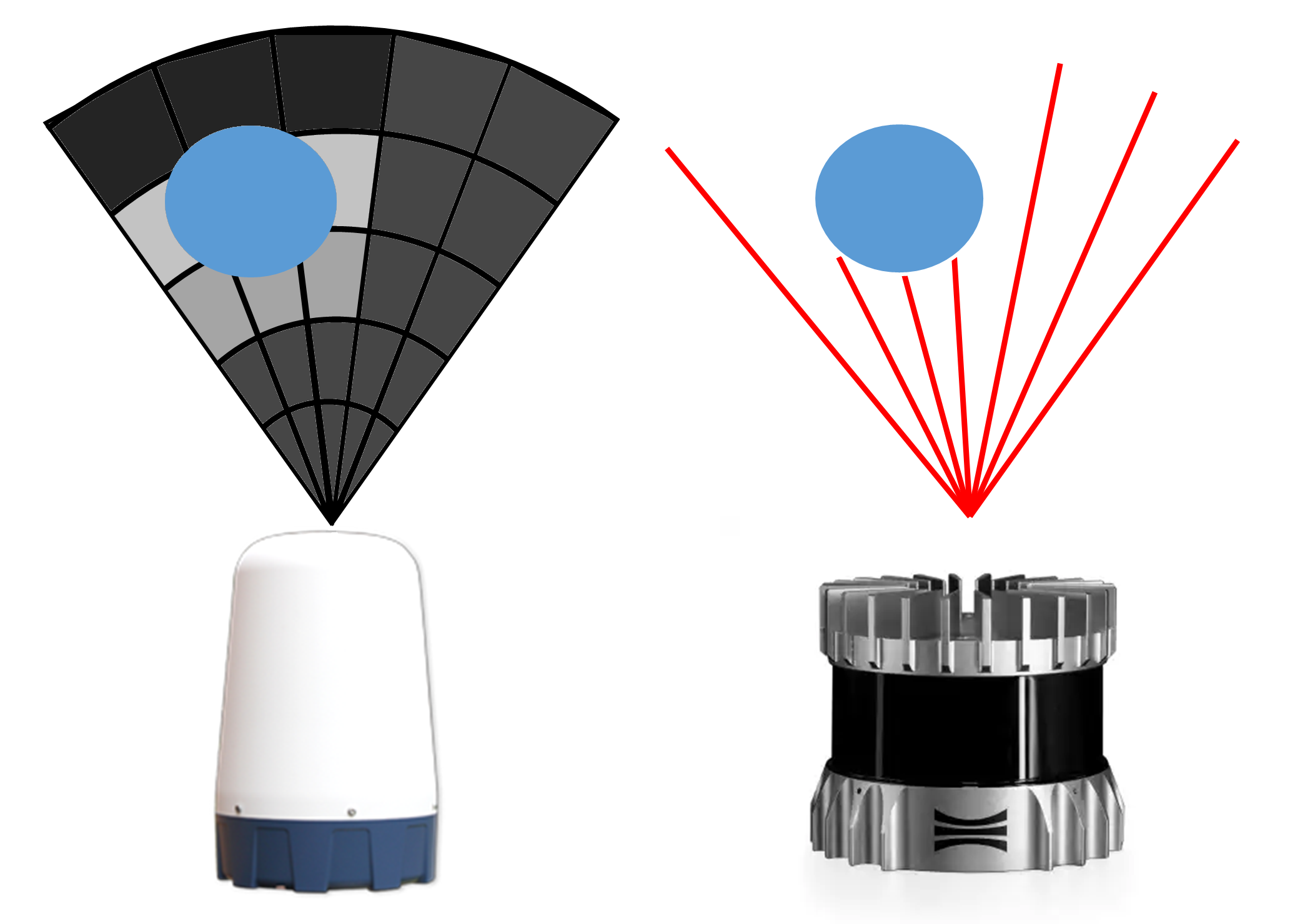}
      \includegraphics[width=\linewidth]{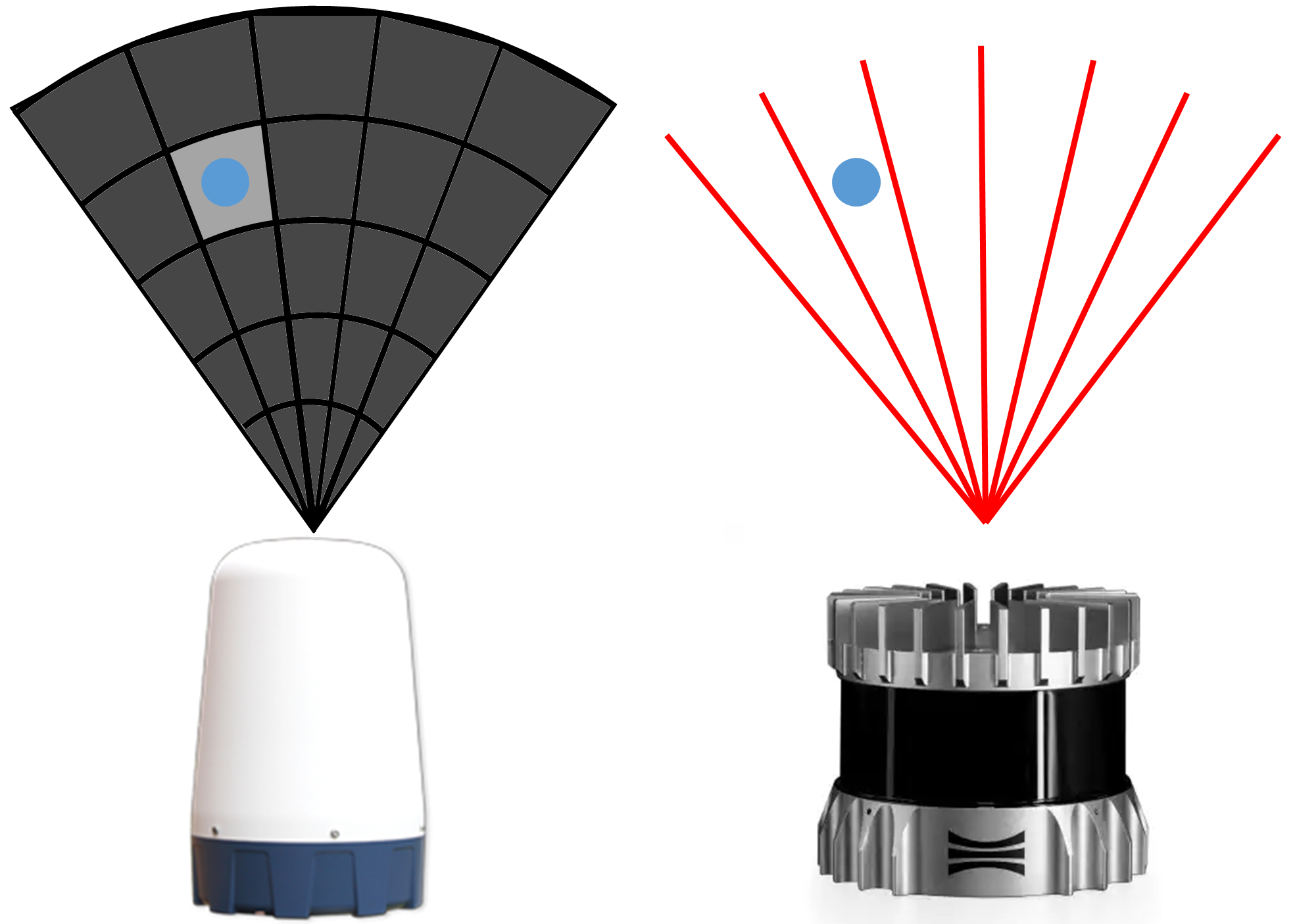}
	}}
      
      \caption{A figure showing the difference between a radar cone (left) and a lidar beam (right) and how it sees objects (blue circle). Shown with a large object on top and a small object on bottom.}
      \label{fig:beam_vs_cone}
   \end{figure}

Lidar beams can be thought of as rays with very little width. These rays, although they exhibit a slight divergence, do so minimally, with just 0.18$^{\circ}$ in our case. This also means that, at long ranges or with small objects, where lidar might not receive any returns radar will still see it, although the return intensity will be lower for smaller objects. Additionally, due to the ability of radar waves to "see through" objects, multiple returns will be picked up for each emitted beam. Fig.~\ref{fig:beam_vs_cone} shows how this works for both sensors. On the left is a radar looking at the blue circle, five beams are shown and each beam is broken up into five range bins. For each range bin the radar reports the intensity of the reflected beam. The two beams on the right that don't intersect the object all have low returns. The six bins that intersect the object provide a higher return. And finally, the three bins behind the object provide a lower return, effectively the shadow of the object. The top shows this with larger objects where lidar receives multiple returns and the object covers multiple radar bins. The bottom shows a small object where lidar receives no returns but the object still occupies part of a radar bin. 

This does mean that for every scan of the radar every range bin for every scan will return intensity data between 0.0 and 1.0. However, most of this data will be nothing but background noise. So before any mapping is done a threshold is applied to the radar data. For this work we used a threshold of 0.26 for autonomous operation with any returns below the threshold discarded. This threshold does allow some false positive returns however we also provide some analysis with a threshold of 0.31. This higher threshold reduces the false positive values but also produces some false negatives.

Our system uses an Ouster OS1-128 lidar sensor with a horizontal resolution of 1024 azimuths and a scanning rate of 20~hz and a Navtech CIR-DEV-X radar sensor with a horizontal resolution of 400 azimuths, a scanning rate of 4~hz, and a range bins of 0.044~m mounted on a Clearpath Robotics Warthog (Fig.~\ref{fig:sensors}). Note that while the Ouster lidar provides a 3D scan, the Navtech radar only provides a 2D scan. No pre-processing was applied to the lidar data. For the radar data, we applied a manually set threshold to filter out noise and used odometry information to compensate for the vehicle's motion. 

\subsection{Mapping}

   \begin{figure}[htbp]
      \centering
      \framebox{\parbox{3in}{
      
      \includegraphics[width=\linewidth]{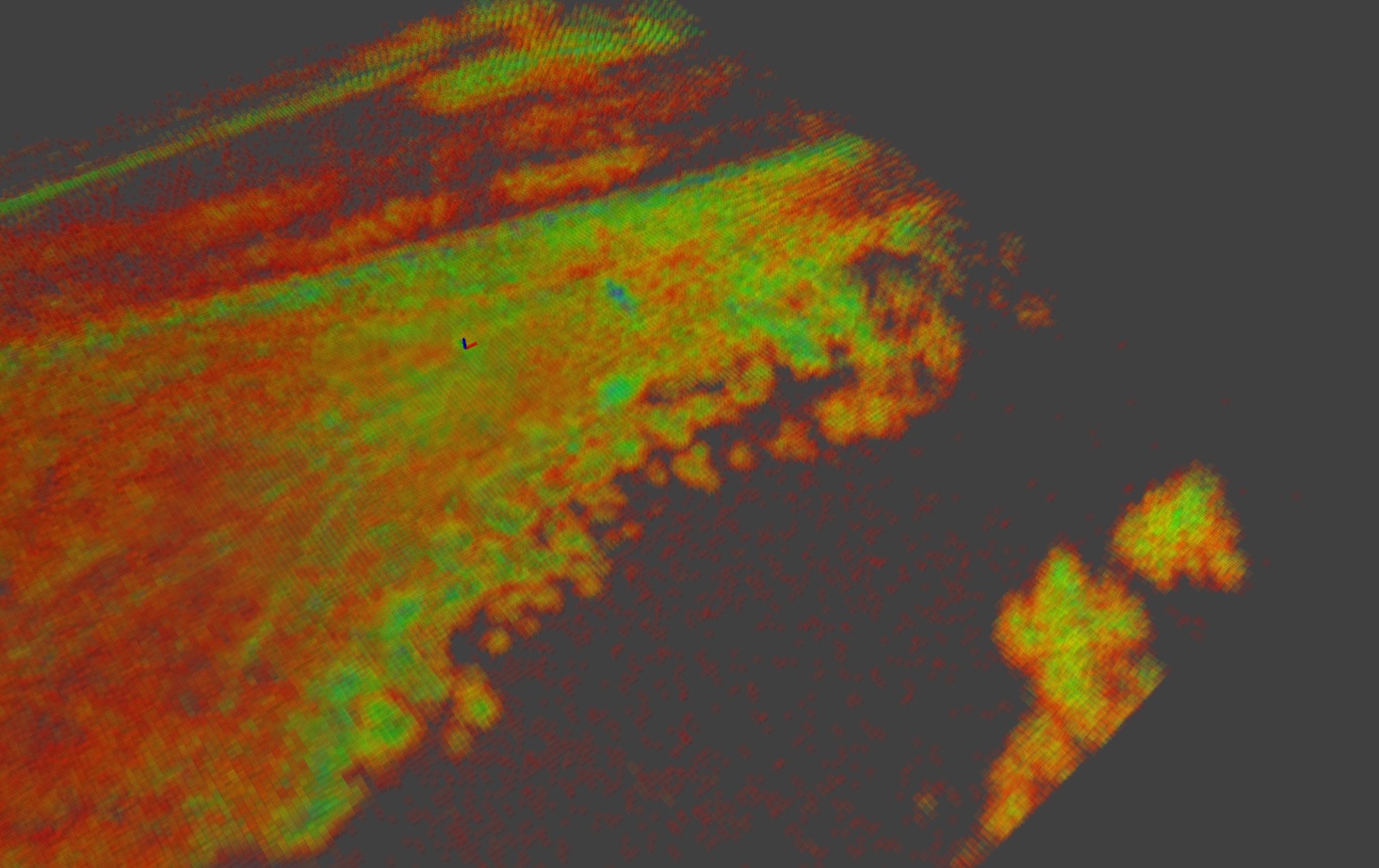}
	}}
      
      \caption{An example of the voxel return density map. Each voxel is colored with the average radar return intensity for that voxels, red is low intensity, green is medium, and blue is high. Voxels are also displayed with 90\% transparency. }
      \label{fig:radar_density}
   \end{figure}


We adapted the GPU-accelerated Voxel Local Mapping (G-VOM) system~\cite{g-vom}, originally developed for lidar-based mapping, to work with radar data. The primary modifications to G-VOM relate to the additional importance of intensity information and ground plane estimation. 

First, while G-VOM already stores the number of hits per voxel, in our modification it also stores the average radar intensity for each voxel. This intensity is used to determine if a voxel is solid or not. Voxels with an average intensity above a set threshold are assumed to be solid, while voxels below the threshold are considered passable. Fig.~\ref{fig:radar_density} shows an example of the average intensity map. 

Next is the ground plane estimation, for each vertical column in the voxel map a weighted average of voxel heights is taken. The weighting for each voxel is the product of the average intensity and the number of hits for that voxel. This approach accounts for the unique characteristics of radar data and enables more accurate ground estimation. 

\begin{figure}[htbp]
      \centering
      \framebox{\parbox{3in}{
      
      \includegraphics[width=\linewidth]{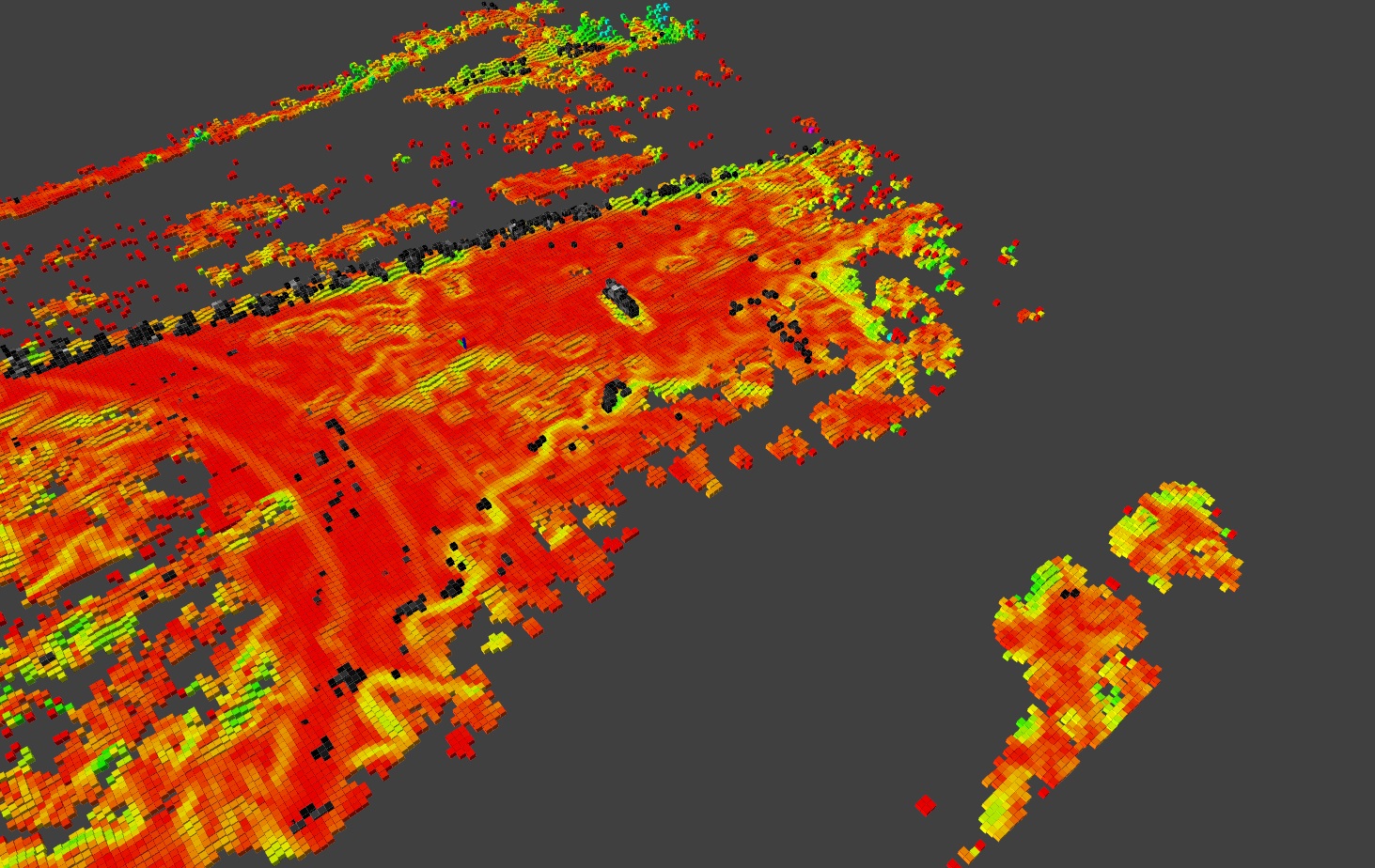}
	}}
      
      \caption{An example of the voxel elevation map and obstacle map. Black voxels are obstacles, colored voxels are the elevation map with red showing low slope and yellow showing high slope. }
      \label{fig:elevation_map}
   \end{figure}

Finally we need to determine where obstacles are. Solid voxels above the ground by an obstacle height threshold are classified as obstacles to be avoided during navigation. Fig.~\ref{fig:elevation_map} shows both the resulting elevation map and obstacle voxels that are derived from the map in Fig.~\ref{fig:radar_density}. 

Once all these terrain metrics have been computed they are used to generate several output maps. First a validity map that indicates whether there is valid data in each cell. Next a slope map for valid cells. And finally an obstacle map. Each of these maps is assigned a cost and a weight, and the final cost map is obtained by summing the weighted maps. The generation of the output maps and the cost map is discussed in more detail in~\cite{g-vom}.

\section{RESULTS}

\subsection{System Overview}

\begin{figure}[htbp]
      \centering
      \framebox{\parbox{3in}{
      
      \includegraphics[width=\linewidth]{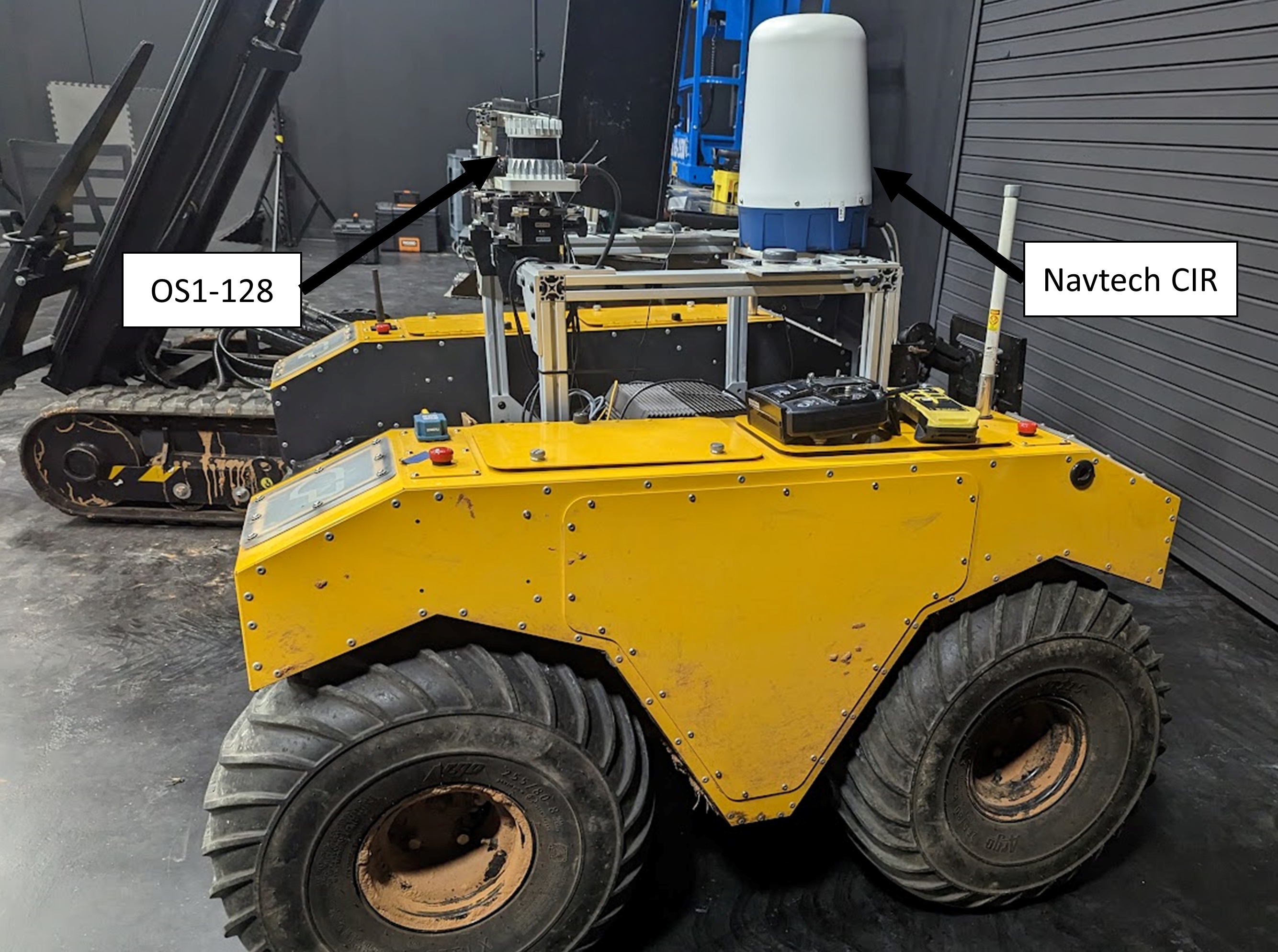}
	}}
      
      \caption{The vehicle, a Clearpath Robotics Warthog, showing the mounting of the lidar, an Ouster OS1-128, and the radar, a Navtech CIR sensor. }
      \label{fig:sensors}
   \end{figure}

\begin{figure}[htbp]
      \centering
      \framebox{\parbox{3in}{
      
      \includegraphics[width=\linewidth]{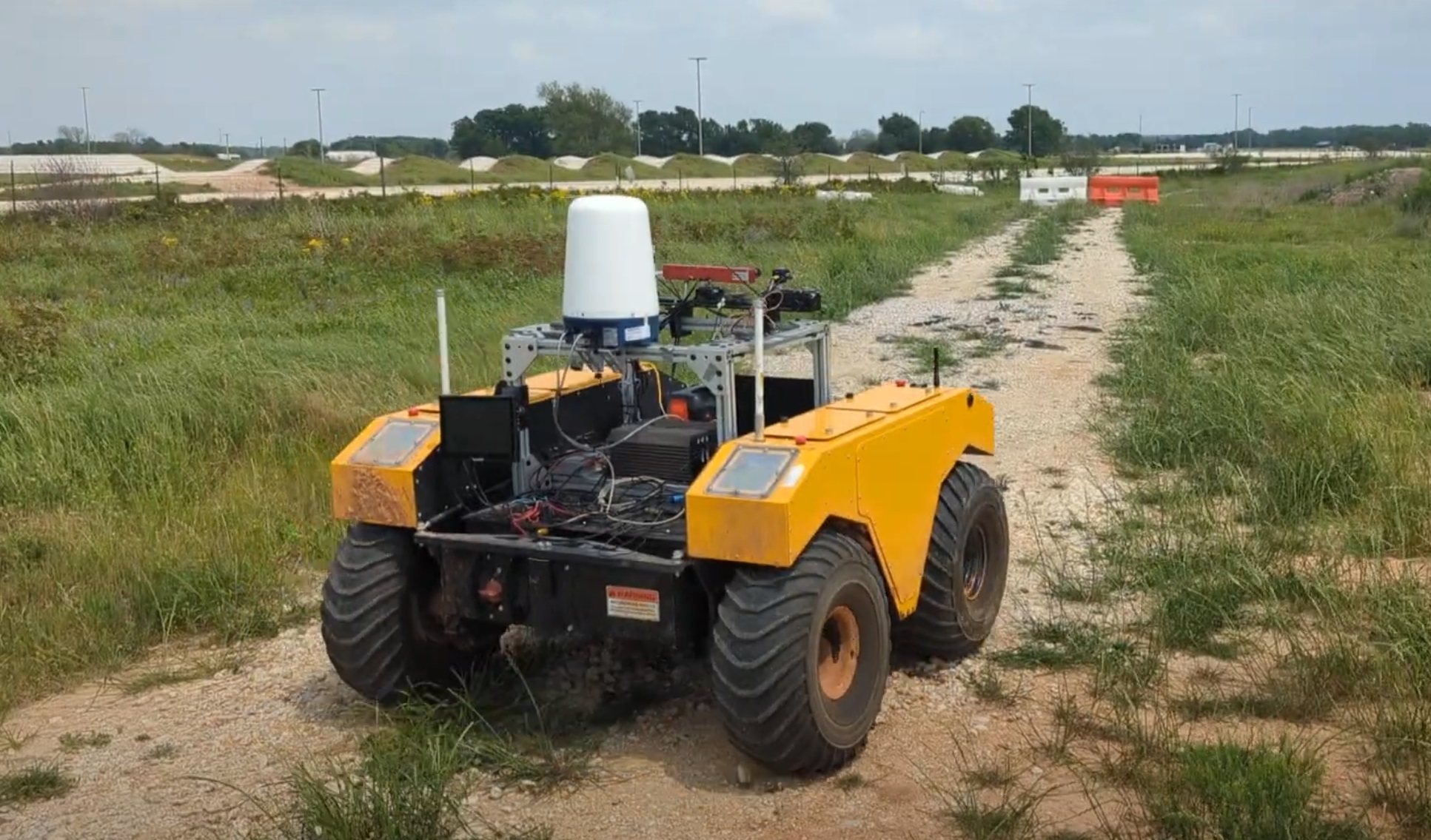}
	}}
      
      \caption{The vehicle at the test site at the beginning of the path. }
      \label{fig:test_site}
   \end{figure}

We implemented our system on a Clearpath Robotics Warthog, shown in Fig.~\ref{fig:sensors}, with an Ouster OS1-128 lidar and a Navtech CIR-DEV-X radar sensor. The lidar was mounted level with the vehicle and the radar was mounted tilted down with a 2.5$^{\circ}$ angle. Since the radar only provides a 2D scan of the environment the motion of the vehicle was used to create the 3D map. The testing environment was the Texas A\&M University RELLIS Campus, depicted in Fig.~\ref{fig:test_site}. We employed Direct lidar Odometry (DLO) \cite{dlo} to supply odometry data. Our adaptation of G-VOM was configured with a voxel resolution of 0.4~m and a map size of 256x256x64. The voxel map was created using only data from the radar using a return intensity threshold of 0.26. Finally, we incorporated the planning methods described in our previous work~\cite{my_planning_paper},\cite{my_optimization_paper}, alongside the controller presented in~\cite{ilqr}. 

\subsection{Autonomous Navigation}

\begin{figure}[htbp]
      \centering
      \framebox{\parbox{3in}{
      
      \includegraphics[width=\linewidth]{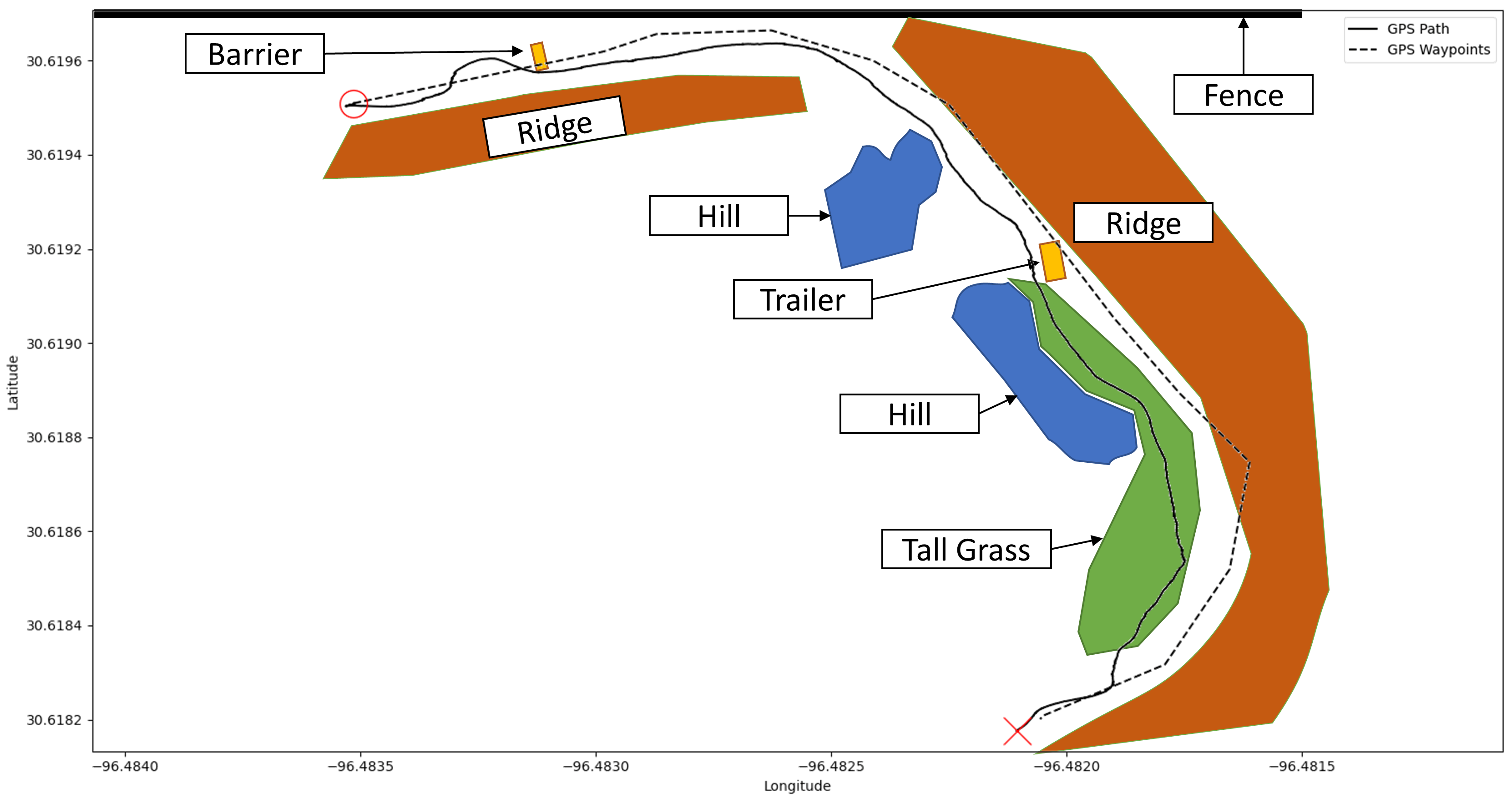}
	}}
      
      \caption{The final driven path overlaid with the gps waypoints. The red circle is the start of the path and the red x is the end. Terrain features have been annotated. }
      \label{fig:competed_path}
   \end{figure}

The data presented in the following sections was collected during autonomous operation of the vehicle. We utilized prerecorded GPS waypoints, spaced 20~m apart, to provide the global path. The vehicle traversed this path at a speed of 2.5~m/s, resulting in a total driven path length of 350~m. Fig.~\ref{fig:competed_path} displays the actual path driven by the vehicle in relation to the given GPS waypoints. The terrain (Fig.~\ref{fig:test_site}) was a mix of grass and gravel with some dense vegetation and solid obstacles, both natural and artificial. The vehicle completed the entire path successfully without requiring any manual interventions. 
   
\subsection{Comparison of Lidar vs Radar Sensor Data}

\begin{figure}[htbp]
      \centering
      \framebox{\parbox{3in}{
      
      \includegraphics[width=\linewidth]{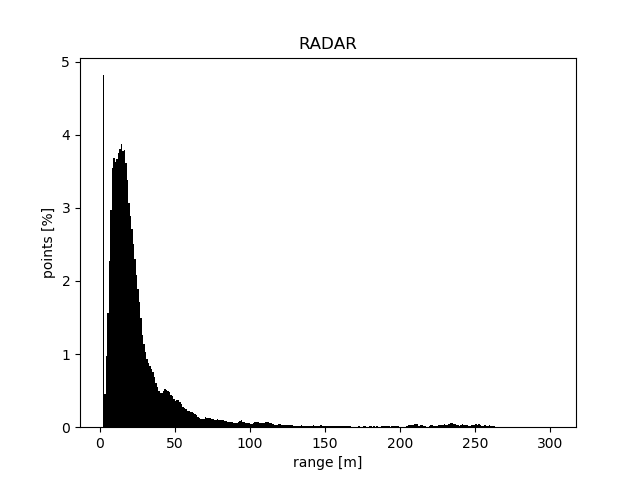}
	}}
      
      \caption{A histogram showing the percentage of radar points returned in each range bin with a threshold of 0.31. }
      \label{fig:radar_hist_20000}
   \end{figure}

In the previous section we showed that it is possible to navigate off-road terrain only using radar data. Now we will discuss the difference between the raw data produced by each sensor and some of the advantages radar provides. The following figures (\ref{fig:lidar_hist},\ref{fig:radar_hist_17000},\ref{fig:radar_hist_20000}) show histograms displaying the percentage of lidar and radar points returned at different ranges. Range bins of 1~m are used for this analysis. As mentioned earlier, the radar data must be thresholded before returns can be interpreted as points. We present histograms at the threshold used for autonomous operation, 0.26, and a higher threshold of 0.31. The higher threshold is presented as the lower threshold does lead to some false positive points. The higher threshold doesn't have these false positives but it was found to work less well during autonomous operation. 

   \begin{figure}[htbp]
      \centering
      \framebox{\parbox{3in}{
      
      \includegraphics[width=\linewidth]{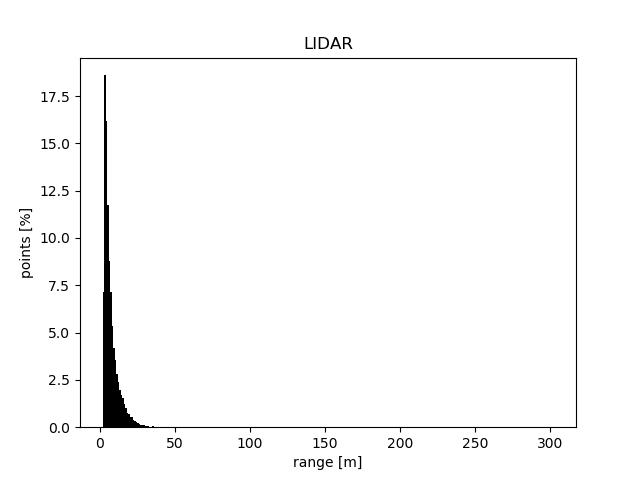}
	}}
      
      \caption{A histogram showing the percentage of lidar points returned in each range bin. }
      \label{fig:lidar_hist}
   \end{figure}
   
   \begin{figure}[htbp]
      \centering
      \framebox{\parbox{3in}{
      
      \includegraphics[width=\linewidth]{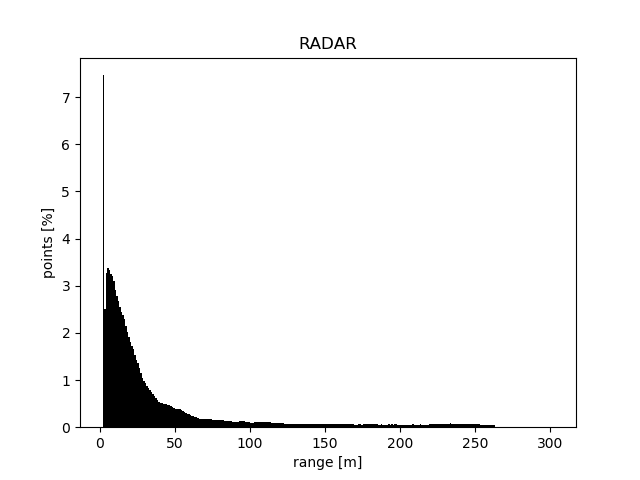}
	}}
      
      \caption{A histogram showing the percentage of radar points returned in each range bin with a threshold of 0.26. }
      \label{fig:radar_hist_17000}
   \end{figure}
   
\begin{figure}[htbp]
      \centering
      \framebox{\parbox{3in}{
      
      \includegraphics[width=\linewidth]{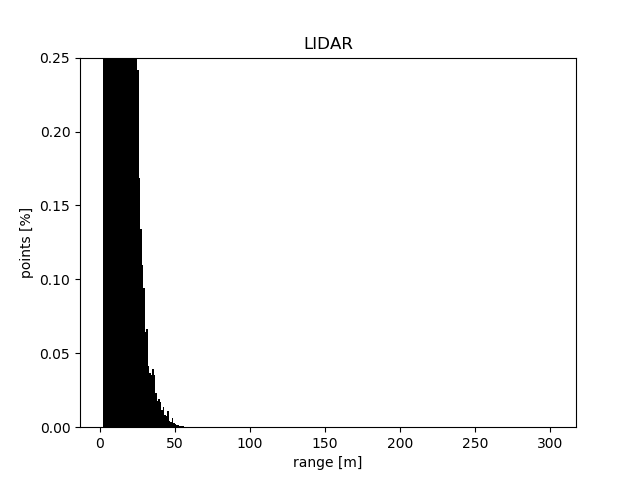}
	}}
      
      \caption{A histogram showing the percentage of lidar points returned in each range bin. Zoomed to show a bin height of 0.25\% and below. }
      \label{fig:lidar_hist_zoom}
   \end{figure}

\begin{figure}[htbp]
      \centering
      \framebox{\parbox{3in}{
      
      \includegraphics[width=\linewidth]{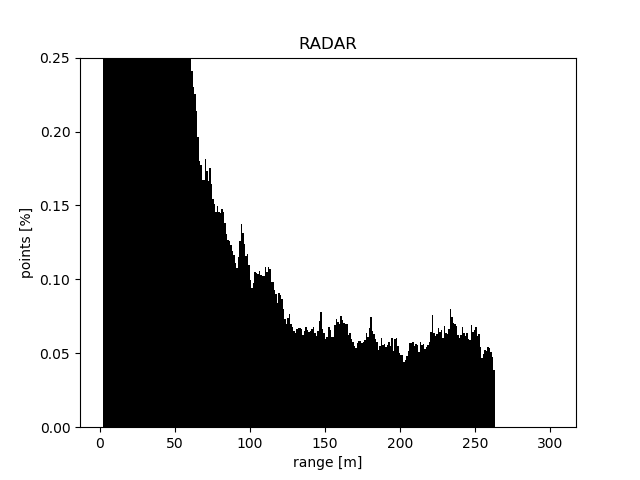}
	}}
      
      \caption{A histogram showing the percentage of radar points returned in each range bin with a threshold of 0.26. Zoomed to show a bin height of 0.25\% and below. }
      \label{fig:radar_hist_zoom_17000}
   \end{figure}

\begin{figure}[htbp]
      \centering
      \framebox{\parbox{3in}{
      
      \includegraphics[width=\linewidth]{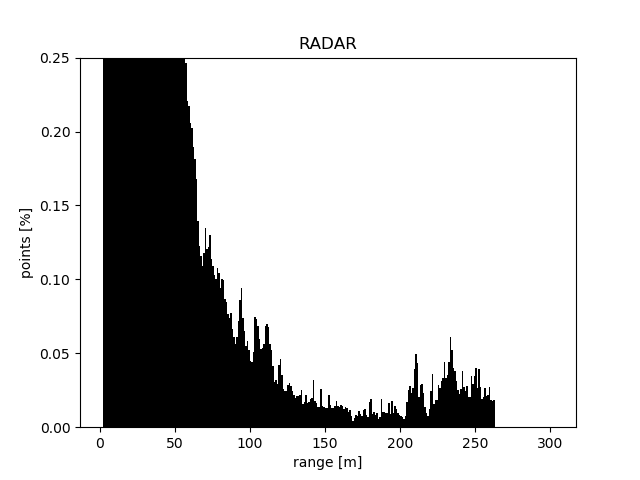}
	}}
      
      \caption{A histogram showing the percentage of radar points returned in each range bin with a threshold of 0.31. Zoomed to show a bin height of 0.25\% and below. }
      \label{fig:radar_hist_zoom_20000}
   \end{figure}

Looking at these figures we immediately see that both radar and lidar show a large number of points near the vehicle with an approximately exponential decay in the number of points as range increases. It is also clear that the radar has a significantly longer range than the lidar. With lidar points having a maximum range of approximately 50~m where the radar still has points beyond 250~m. Figures \ref{fig:lidar_hist_zoom},\ref{fig:radar_hist_zoom_17000},and \ref{fig:radar_hist_zoom_20000} show this better by zooming into the tails of each histogram, showing a upper bin size of 0.25\%. 

We can see that the lidar has a bin size of 0.25\% at approximately 25~m (Fig.~\ref{fig:lidar_hist_zoom}), the lower radar threshold shows the same bin size at 55~m (Fig.~\ref{fig:radar_hist_zoom_17000}), more than twice the range of the lidar. Even at the higher threshold we still see the radar having a 0.25\% range of over 50~m (Fig.~\ref{fig:radar_hist_zoom_20000}). Looking at the maximum range the difference is even greater, with just above 50~m for the lidar and over 250~m for the radar at both threshold values. That means the radar has a maximum range over five times that of lidar. 

\begin{figure}[htbp]
      \centering
      \framebox{\parbox{3in}{
      
      \includegraphics[width=\linewidth]{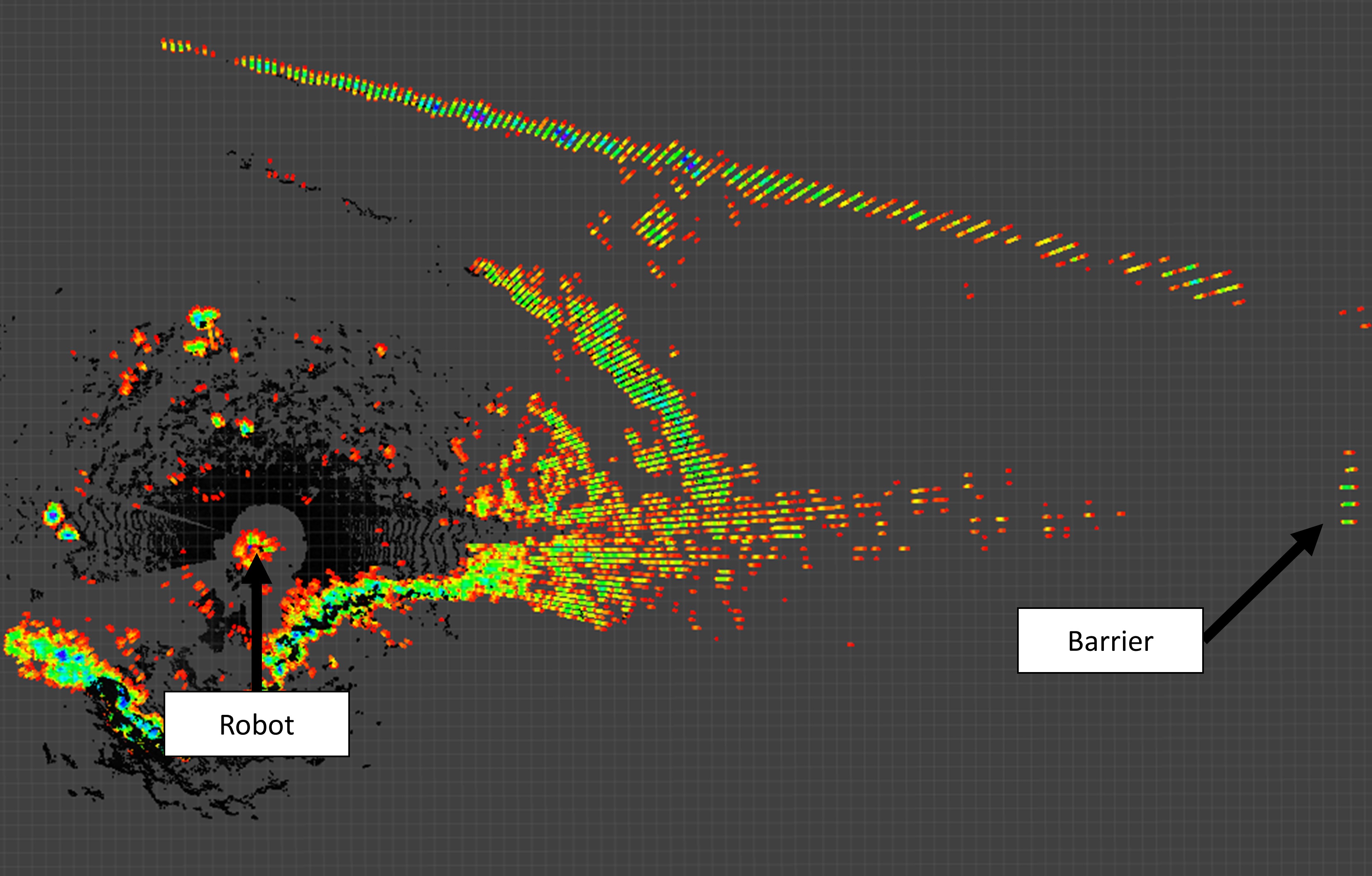}
	}}
      
      \caption{A top down view of lidar data (black) and radar data (colored) showing detection of the barrier in radar at 80~m. Note that the farthest lidar points are only at 40~m. }
      \label{fig:obs_raw}
   \end{figure}

But how does this compare when looking at a real object? Fig.~\ref{fig:obs_raw} shows both a radar and lidar view from where the vehicle is in Fig.~\ref{fig:test_site}. This is at the maximum range the vehicle is able to detect the barrier in lidar. We found that the barrier could be easily seen by the radar at a range of 80~m whereas it was not visible to the lidar until the vehicle was within 40~m. This matches well with our 0.25\% range, showing the radar with twice the effective range of the lidar. 
   
\subsection{Comparison of Lidar vs Radar Maps}

\begin{figure}[htbp]
      \centering
      \framebox{\parbox{3in}{
      
      \includegraphics[width=\linewidth]{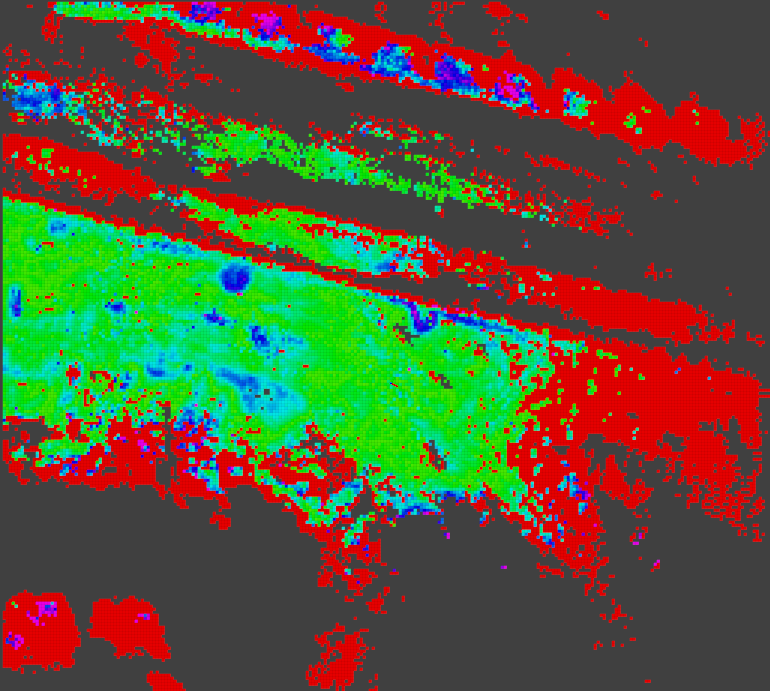}
	}}
      
      \caption{A figure showing the difference between the lidar and radar height maps. The red region is where there is radar data but no lidar data. Pink regions are greater than 1~m error. Green through blue is 0~m to 1~m error. }
      \label{fig:lidar_radar_maps}
   \end{figure}

\begin{figure}[htbp]
      \centering
      \framebox{\parbox{3in}{
      
      \includegraphics[width=\linewidth]{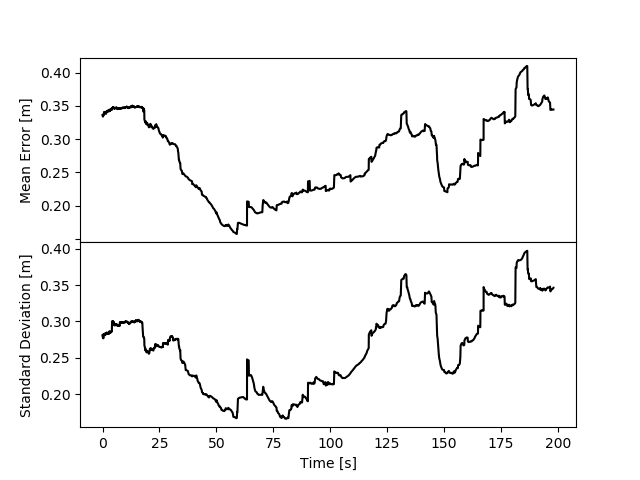}
	}}
      
      \caption{A figure showing the error and standard deviation over time between the lidar and radar height maps. }
      \label{fig:lidar_radar_maps_plot}
   \end{figure}

In the previous section we showed that the radar can see significantly farther than the lidar. Now we compare the radar based height map with the more traditional lidar based maps using the lidar maps as the ground truth. Fig.~\ref{fig:lidar_radar_maps} shows a top down view of the difference between the radar and lidar height maps. Note that not only does the resulting radar map have a noticeably larger range but the error, calculated as the absolute difference between the radar and lidar height map, is also very low. With only small regions with an error greater than 1~m. But this is only one frame of data, Fig.~\ref{fig:lidar_radar_maps_plot} shows the mean error and standard deviation of the error over the entire path length. We can see that the maximum mean error over the entire path is only 0.4~m with the majority of the error being less than 0.3~m. The standard deviation follows a similar trend with the maximum also being 0.4~m and most being under 0.3~m. Looking again at Fig.~\ref{fig:lidar_radar_maps} we can see that this large standard deviation is easily explained by large regions of the height map near the vehicle having a lower error with smaller regions far from the vehicle having a much larger error. We found that, since the elevation mapping relies on the average of many scans, the resulting elevation map becomes more accurate the closer it is to the vehicle. 

\begin{figure}[htbp]
      \centering
      \framebox{\parbox{3in}{
      
      \includegraphics[width=\linewidth]{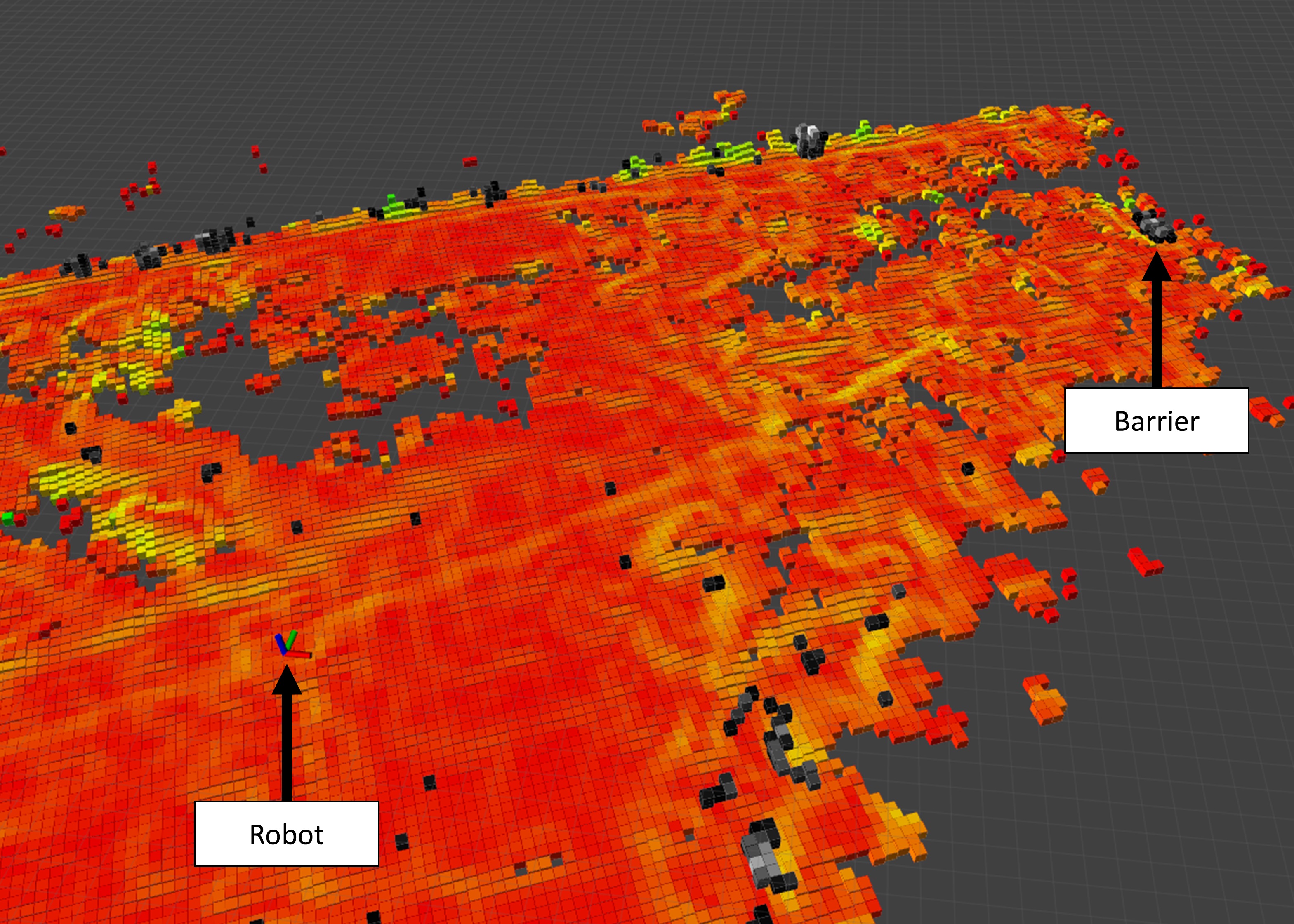}
	}}
      
      \caption{A figure showing the distance that an object is seen in the radar map. 50~m }
      \label{fig:obs}
   \end{figure}

That is good for elevation but we are also interested in obstacle detection. It is harder to quantitatively define the quality of obstacle maps. However, as with the raw sensor data, the radar appears to detect positive obstacles at a range of at least 50~m as shown in Fig.~\ref{fig:obs}. This is right at the edge of the map given the map has a side length of 102.4~m and the vehicle is at the center of the map. Qualitatively, successful autonomous operation without collisions over a path length of 350~m, as described above, demonstrates that the obstacle detection is sufficient.

\section{CONCLUSIONS}

In this paper, we have presented an evaluation of radar alone as a potential alternative to lidar for off-road local navigation. By adapting existing lidar-based methods for use with radar, we have demonstrated the feasibility of radar only navigation in off-road conditions. We show the successful navigation of an off-road ground vehicle using only radar for obstacle detection and terrain estimation. Additionally, we show that radar has a significantly longer range than lidar, offering more than twice the effective detection range. This increased range was apparent in both histogram analysis and real-world object detection. Even despite radar's lower resolution, the generated maps based on radar data exhibited a similar level of accuracy as those created from lidar data. 
Finally, our findings support the notion that radar can serve as a viable alternative or complement to traditional lidar-based approaches, ultimately enhancing the robustness and reliability of navigation systems in off-road environments. 

\balance




\bibliographystyle{IEEEtran}
\bibliography{IEEEabrv,IEEEexample}

\begin{thebibliography}{10}
\providecommand{\url}[1]{#1}
\csname url@rmstyle\endcsname
\providecommand{\newblock}{\relax}
\providecommand{\bibinfo}[2]{#2}
\providecommand\BIBentrySTDinterwordspacing{\spaceskip=0pt\relax}
\providecommand\BIBentryALTinterwordstretchfactor{4}
\providecommand\BIBentryALTinterwordspacing{\spaceskip=\fontdimen2\font plus
\BIBentryALTinterwordstretchfactor\fontdimen3\font minus
  \fontdimen4\font\relax}
\providecommand\BIBforeignlanguage[2]{{%
\expandafter\ifx\csname l@#1\endcsname\relax
\typeout{** WARNING: IEEEtran.bst: No hyphenation pattern has been}%
\typeout{** loaded for the language `#1'. Using the pattern for}%
\typeout{** the default language instead.}%
\else
\language=\csname l@#1\endcsname
\fi
#2}}

\bibitem{radar_slam_2013}
\BIBentryALTinterwordspacing
D.~Vivet, F.~Gérossier, P.~Checchin, L.~Trassoudaine, and R.~Chapuis, ``Mobile
  ground-based radar sensor for localization and mapping: An evaluation of two
  approaches,'' \emph{International Journal of Advanced Robotic Systems},
  vol.~10, no.~8, p. 307, 2013. [Online]. Available:
  \url{https://doi.org/10.5772/56636}
\BIBentrySTDinterwordspacing

\bibitem{radar_lidar_slam_2019}
M.~Mielle, M.~Magnusson, and A.~J. Lilienthal, ``A comparative analysis of
  radar and lidar sensing for localization and mapping,'' in \emph{2019
  European Conference on Mobile Robots (ECMR)}, 2019, pp. 1--6.

\bibitem{radar_vs_lidar_2009}
\BIBentryALTinterwordspacing
J.~Ryde and N.~Hillier, ``Performance of laser and radar ranging devices in
  adverse environmental conditions,'' \emph{Journal of Field Robotics},
  vol.~26, no.~9, pp. 712--727, 2009. [Online]. Available:
  \url{https://onlinelibrary.wiley.com/doi/abs/10.1002/rob.20310}
\BIBentrySTDinterwordspacing

\bibitem{radar_lidar_slam_2018}
\BIBentryALTinterwordspacing
P.~Fritsche, S.~Kueppers, G.~Briese, and B.~Wagner, \emph{Fusing LiDAR and
  Radar Data to Perform SLAM in Harsh Environments}.\hskip 1em plus 0.5em minus
  0.4em\relax Cham: Springer International Publishing, 2018, pp. 175--189.
  [Online]. Available: \url{https://doi.org/10.1007/978-3-319-55011-4\_9}
\BIBentrySTDinterwordspacing

\bibitem{radar_slam_2022}
K.~Burnett, Y.~Wu, D.~J. Yoon, A.~P. Schoellig, and T.~D. Barfoot, ``Are we
  ready for radar to replace lidar in all-weather mapping and localization?''
  \emph{IEEE Robotics and Automation Letters}, vol.~7, no.~4, pp.
  10\,328--10\,335, 2022.

\bibitem{radar_on_lidar_slam_2020}
H.~Yin, Y.~Wang, L.~Tang, and R.~Xiong, ``Radar-on-lidar: metric radar
  localization on prior lidar maps,'' in \emph{2020 IEEE International
  Conference on Real-time Computing and Robotics (RCAR)}, 2020, pp. 1--7.

\bibitem{radar_on_lidar_slam_2022}
H.~Yin, R.~Chen, Y.~Wang, and R.~Xiong, ``Rall: End-to-end radar localization
  on lidar map using differentiable measurement model,'' \emph{IEEE
  Transactions on Intelligent Transportation Systems}, vol.~23, no.~7, pp.
  6737--6750, 2022.

\bibitem{radar_lidar_dust_2009}
T.~Peynot, J.~Underwood, and S.~Scheding, ``Towards reliable perception for
  unmanned ground vehicles in challenging conditions,'' in \emph{2009 IEEE/RSJ
  International Conference on Intelligent Robots and Systems}, 2009, pp.
  1170--1176.

\bibitem{radar_perception_2011}
\BIBentryALTinterwordspacing
G.~Reina, J.~Underwood, G.~Brooker, and H.~Durrant-Whyte, ``Radar-based
  perception for autonomous outdoor vehicles,'' \emph{Journal of Field
  Robotics}, vol.~28, no.~6, pp. 894--913, 2011. [Online]. Available:
  \url{https://onlinelibrary.wiley.com/doi/abs/10.1002/rob.20393}
\BIBentrySTDinterwordspacing

\bibitem{imaging_radar_2019}
D.~Gusland, B.~Torvik, E.~Finden, F.~Gulbrandsen, and R.~Smestad, ``Imaging
  radar for navigation and surveillance on an autonomous unmanned ground
  vehicle capable of detecting obstacles obscured by vegetation,'' in
  \emph{2019 IEEE Radar Conference (RadarConf)}, 2019, pp. 1--6.

\bibitem{radar_vision_obs_2015}
G.~Reina, A.~Milella, and R.~Rouveure, ``Traversability analysis for off-road
  vehicles using stereo and radar data,'' in \emph{2015 IEEE International
  Conference on Industrial Technology (ICIT)}, 2015, pp. 540--546.

\bibitem{radar_lidar_dust_2012}
M.~P. Gerardo-Castro and T.~Peynot, ``Laser-to-radar sensing redundancy for
  resilient perception in adverse environmental conditions,'' 01 2012.

\bibitem{radar_lidar_obs_2015}
\BIBentryALTinterwordspacing
J.~Hollinger, B.~Kutscher, and R.~Close, ``{Fusion of lidar and radar for
  detection of partially obscured objects},'' in \emph{Unmanned Systems
  Technology XVII}, R.~E. Karlsen, D.~W. Gage, C.~M. Shoemaker, and G.~R.
  Gerhart, Eds., vol. 9468, International Society for Optics and
  Photonics.\hskip 1em plus 0.5em minus 0.4em\relax SPIE, 2015, p. 946806.
  [Online]. Available: \url{https://doi.org/10.1117/12.2177050}
\BIBentrySTDinterwordspacing

\bibitem{g-vom}
T.~Overbye and S.~Saripalli, ``G-vom: A gpu accelerated voxel off-road mapping
  system,'' in \emph{2022 IEEE Intelligent Vehicles Symposium (IV)}, 2022, pp.
  1480--1486.

\bibitem{dlo}
K.~Chen, B.~T. Lopez, A.-a. Agha-mohammadi, and A.~Mehta, ``Direct lidar
  odometry: Fast localization with dense point clouds,'' \emph{IEEE Robotics
  and Automation Letters}, vol.~7, no.~2, pp. 2000--2007, 2022.

\bibitem{my_planning_paper}
T.~Overbye and S.~Saripalli, ``Fast local planning and mapping in unknown
  off-road terrain,'' in \emph{2020 IEEE International Conference on Robotics
  and Automation (ICRA)}, 2020, pp. 5912--5918.

\bibitem{my_optimization_paper}
------, ``Path optimization for ground vehicles in off-road terrain,'' in
  \emph{2021 IEEE International Conference on Robotics and Automation (ICRA)},
  2021.

\bibitem{ilqr}
A.~Nagariya and S.~Saripalli, ``An iterative lqr controller for off-road and
  on-road vehicles using a neural network dynamics model,'' in \emph{2020 IEEE
  Intelligent Vehicles Symposium (IV)}, 2020, pp. 1740--1745.

\end{thebibliography}

\end{document}